\documentclass{article}

 \usepackage[preprint]{neurips_2026}

\usepackage[utf8]{inputenc} 
\usepackage{amsmath}
\usepackage[T1]{fontenc}    
\usepackage{hyperref}       
\usepackage{url}            
\usepackage{booktabs}       
\usepackage{amsfonts}       
\usepackage{nicefrac}       
\usepackage{microtype}      
\usepackage{xcolor}         
\usepackage{graphicx}       

\title{Structural Hierarchy and Geometry in Molecular Representation Learning}

\author{
  \parbox{\textwidth}{\centering
    \textbf{David Sulu\thanks{Equal contribution}} \quad 
    \textbf{Lorenzo Di Fruscia\footnotemark[1]} \quad 
    \textbf{Jana M. Weber\thanks{Corresponding author}} \\[6pt]
    {\normalfont Department of Intelligent Systems, Delft University of Technology, Delft, The Netherlands} \\
    {\small \texttt{\{l.difruscia, j.m.weber\}@tudelft.nl}}
  }
}

\begin{document}

\maketitle

\begin{abstract}
Molecular self-supervised learning uses chemical structures to guide which molecular embeddings should be similar. We study whether explicitly encoding a molecule’s Bemis–Murcko scaffold and using it to supervise the molecular embedding changes what the model learns. We further test whether this effect depends on the embedding geometry by comparing Euclidean and Lorentz contrastive objectives. Across two augmentation strengths, scaffold-supervised models consistently organize molecules according to both identical and structurally related scaffolds. The resulting embeddings also improve molecular property prediction on several tasks, while the exact gains depend on the predicted property. The effect of scaffold supervision on molecular organization is stronger under Lorentz objectives, but neither geometry provides a consistent overall advantage. These results show that explicitly teaching the relation between a molecule and its structural core can reliably shape the organization of molecular embedding space, while the extent of usefulness of this organization remains task dependent.
\end{abstract}

\section{Introduction}

Functional molecular properties have been measured for only a small fraction of molecules in chemical space.
Experimental validation is costly, and this limits the amount of labelled data available for training molecular property prediction models.
Self-Supervised Learning (SSL) is a family of strategies that help mitigate this issue: by learning reusable molecular embeddings from large collections of unlabeled structures, it can improve predictive performance on tasks where labelled data are scarce \citep{10.1093/nar/gky1033,MoleculeNet}.

Contrastive learning is one widely used SSL strategy. By pulling together related views (positives) while separating other instances (negatives), it teaches the 
model to learn an embedding from a notion of similarity \citep{SimCLR,InfoNCE}. For example, methods such as GraphCL and MolCLR construct two perturbed versions of the same molecular graph by masking atom information, altering bonds or removing whole subgraphs, and train the model to align their embeddings \citep{GraphContrastiveLearningWithAugmentations}, \citep{MOLCLR}. Subsequent work makes molecular contrastive learning more chemically informed in several ways. Fragment-based approaches decompose molecules into smaller subgraphs: iMolCLR augments molecule-level contrastive learning with a fragment-level contrastive objective and reduces the repulsion of chemically similar molecular negatives \citep{iMolCLR}; GraphFP learns across molecular and fragment resolutions by contrasting fragment embeddings with embeddings of their corresponding atoms \citep{2023fragmentbased}; HiMol uses selected chemically meaningful fragments, termed motifs, and incorporates them as an intermediate level in an atom--motif--molecule hierarchy \citep{HierarchicalMolecularGraphSelf}.
Scaffold-aware approaches instead use information about the molecular core. \citep{Wan2025} use scaffold information to construct scaffold-preserving molecular views and to modulate how strongly molecules with different scaffolds are separated during contrastive learning.

The geometry of the embedding space may also affect how relations between structural levels are represented.
Standard contrastive learning typically compares embeddings in Euclidean space. 
Hierarchical or branching relationships in data are efficiently, with low distortion, represented in hyperbolic spaces \citep{Poincare,LorentzEmbeddings,Sarkar2011, MERU}. Molecular structure can be described at multiple levels of abstraction, but this does not imply that molecular space itself must be hierarchical or inherently hyperbolic. Previous molecular work has explored hyperbolic geometry using an externally supplied hierarchy, namely a therapeutic taxonomy, rather than inducing the hierarchy through the pretraining objective itself \citep{Yu2022Hyperbolic}.

Molecular cores are used to organize related compound series and can capture chemically meaningful common structures \citep{BemisMurcko,Wan2025}. They thus provide a natural coarser structural abstraction. The effect of using such a shared abstraction as a direct target for whole-molecule embeddings is worth studying because the same core is a common target for many compounds, so the objective introduces an explicit cross-level, many-to-one relation rather than the one-to-one positive pairing used in standard contrastive learning. This has not been tested in the fragment- and scaffold-based approaches above. Scaffold supervision provides a minimal way of introducing scaffold-level structural organization while leaving the molecular encoder and augmentation scheme unchanged, allowing the effect of the additional supervision to be studied in isolation.

This work investigates the effect of using Bemis--Murcko (BM) scaffolds as an explicit many-to-one supervision signal in molecular contrastive learning \citep{BemisMurcko}. Introducing a hierarchical relationship, we thus further test whether the choice of embedding geometry matters. We assess both aspects along two complementary dimensions: molecular neighbourhood organization through molecule--molecule retrieval, and downstream transfer through frozen molecular property prediction. 

\section{Methods}

\textbf{Molecular representation and embedding model.}\quad
Each molecule is represented as a graph $G=(V,E)$, with atoms as nodes and bonds as edges. A Graph Isomorphism Network with edge features (GINE) encodes this graph into a molecular embedding $h=f_\theta(G)$. We use $h$ for molecule retrieval and property prediction. During contrastive pretraining, a projection head maps $h$ to a second embedding $z=g_\phi(h)$, where the training loss is applied \citep{GIN,PretrainGNN}.
The contrastive objective acts on $z$, so the projection head mediates this objective, while $h$ is retained as the reusable molecular embedding for downstream evaluation, following standard contrastive-learning practice.

\textbf{Contrastive pretraining objective.}\quad
Following MolCLR \citep{MOLCLR}, for each molecule $G_i$ we create two perturbed views $G_i^{(1)}$ and $G_i^{(2)}$ by 
randomly masking atoms and deleting bonds at the selected perturbation rate. The model is challenged to place the two versions of the same molecule close together and embeddings of different molecules farther apart. We use perturbation strength of 25\%, following MolCLR, and 15\% as a milder augmentation setting.
For computational reasons, all models are pretrained on a 1M subset of the released MolCLR dataset using a 75/25 training-validation split, and each condition is trained independently with three random seeds.
The full loss is given in Appendix Eq.~\ref{eq:ntxent}.

\begin{figure}[h!]
  \centering
  \includegraphics[width=\linewidth]{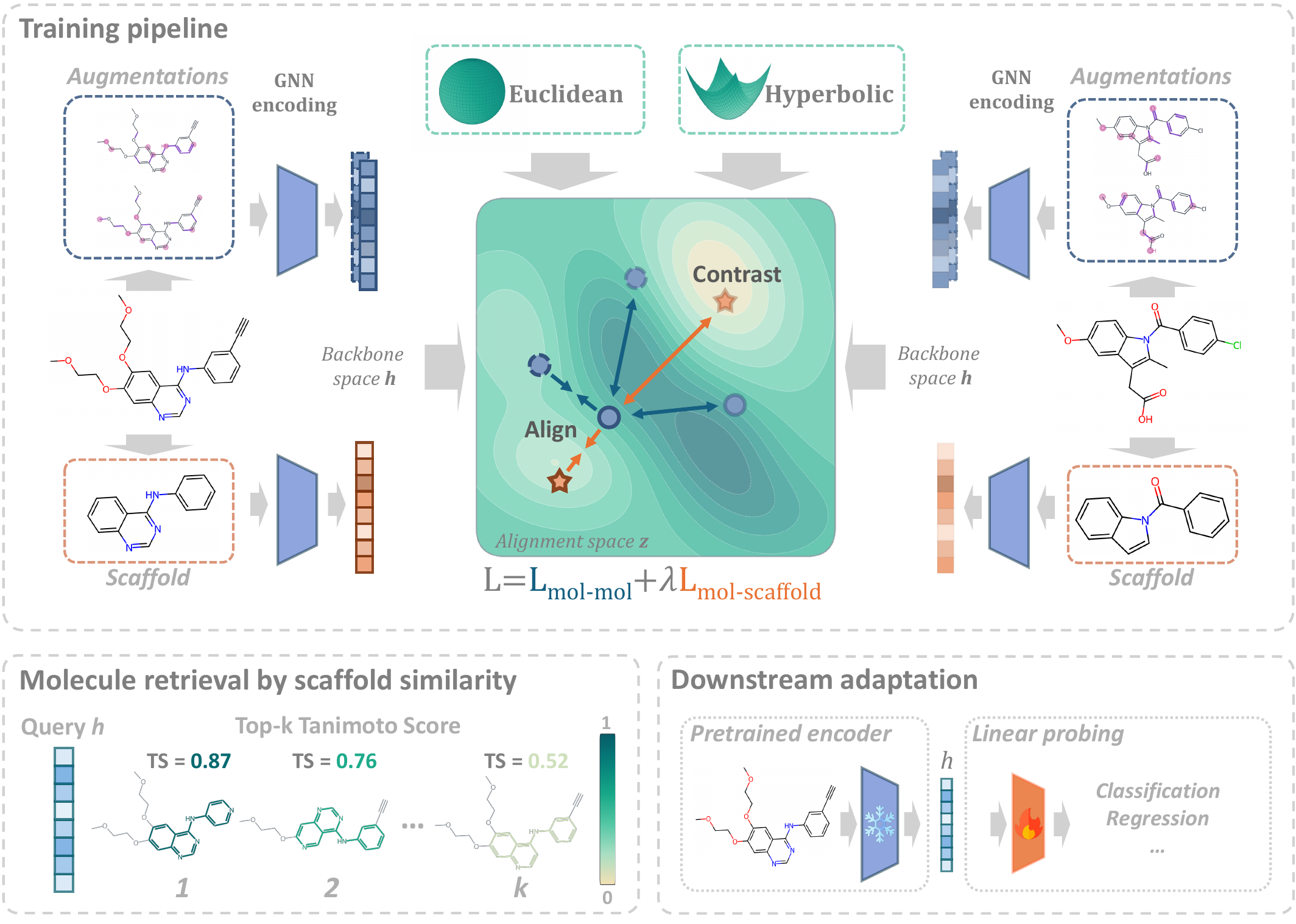}
  \caption{\textit{Top}: Scaffold-aware contrastive pretraining. Augmented molecular views and their
  unaugmented BM scaffolds share an encoder and projection head. Augmented molecular views
  are aligned/contrasted with one another, while explicit scaffold supervision aligns each view
  with its own scaffold and contrasts it against other scaffold embeddings.
  The contrastive objectives are evaluated under either Euclidean or Lorentz projection geometry.
  \textit{Bottom}: After pretraining, the frozen backbone space $h$ is evaluated through molecular
  retrieval and linear property prediction.}
  
  \label{fig:method}
\end{figure}

\textbf{Scaffold supervision.}\quad
We additionally extract the BM scaffold of each molecule, retaining ring systems and their connecting linkers while removing peripheral side chains, and encode it in the same graph encoder. During training, each view is encouraged to be similar to its own scaffold and different from the scaffolds of other molecules in the batch (Appendix Eq.~\ref{eq:scaffold}).
The total loss combines $\mathcal L_{\rm mol}$ and $\mathcal L_{\rm scaff}$ with scaffold weight $\lambda_s=0.1$. Scaffold supervision therefore explicitly teaches the model the relation between a molecule and its structural core. Figure~\ref{fig:method} summarizes the pretraining design.

\textbf{Embedding geometry.}\quad
The Euclidean and Lorentz variants share the same setup and differ only in how similarity between projected embeddings is computed.
In the Euclidean variant, projected vectors $z$ are compared using cosine similarity. In the Lorentz variant, 
$z$ is first mapped from Euclidean space onto a Lorentz hyperboloid, and contrastive similarity is 
defined as the negative geodesic distance between the resulting points.
We use a radially bounded Lorentz formulation throughout the main text. Both the bounded and unbounded formulations are defined in Appendix~\ref{sec:geometry}, and Appendix~\ref{sec:diagnostics} reports the diagnostic that motivates this choice.
The transformation acts only on $z$, while the backbone embedding $h$ remains Euclidean.

\textbf{Evaluation.}\quad
We evaluate the frozen molecular embeddings $h$ in two ways: we study how scaffold supervision reorganizes the molecular embedding space, and we test how well the embeddings transfer knowledge to downstream molecular property prediction. 

To study latent space re-organization, we retrieve molecular neighbours from a fixed cohort of 100,000 unique molecules from the pretraining validation set using cosine similarity in $h$. Note that this is not a general measure of embedding quality, but simply describes the re-organization. We measure two distinct phenomena: the recovery of molecules sharing the query's exact BM scaffold using mean average precision at ten (MAP@10) as a model consistency check, and the ECFP4--Tanimoto similarity (Tan@10) of non-identical scaffolds with its normalized discounted cumulative gain at ten (nDCG@10). The latter quantifies similarity and ranking quality and is done either between the corresponding BM scaffolds (BM-Tan@10, BM-nDCG@10) or between the complete molecules (Mol-Tan@10, Mol-nDCG@10). Full metric definitions, cohort construction, and reference baselines are provided in Appendix~\ref{sec:structural-diagnostics}.

We separately evaluate downstream transfer on seven MoleculeNet tasks, comprising three classification and four regression problems \citep{MoleculeNet}. We report them in Appendix Table ~\ref{tab:moleculenet-tasks}. For each task, we freeze the encoder $f_\theta$ and train a linear predictor on $h$ using a fixed 80/10/10 scaffold split. We report mean $\pm$ SD across three independently pretrained encoders. This evaluation tests whether the learned embeddings contain useful information for property prediction tasks.

\section{Results}

We first study how scaffold supervision changes the organization of the learned molecular embeddings, and then ask whether these changes are useful for selected molecular property prediction tasks.

\textbf{Scaffold supervision reorganizes molecular neighbourhoods.}\quad
As expected, scaffold supervision strongly increases retrieval of molecules sharing the query's exact BM scaffold (Appendix Table~\ref{tab:retrieval-h}). Because exact scaffold identity is used during training, this confirms that the model learns the intended scaffold relation.

We next exclude all molecules sharing the query's exact scaffold. Under this stricter setting, scaffold-supervised models retrieve molecules with more similar, but nonidentical, BM scaffolds than their corresponding MolCLR models (Table~\ref{tab:retrieval-main}). The same retrieved neighbours are also more similar when the complete molecules are compared using ECFP4.
This effect is positive across all pretraining seeds, geometries, and augmentation settings. 
Scaffold supervision therefore changes molecular neighbourhoods beyond simply grouping molecules with the exact scaffold used during training. Backbone-space t-SNE visualizations show the same qualitative pattern under both alignment geometries (Appendix Figures~\ref{fig:embedding-scaffold-clusters} and ~\ref{fig:embedding-scaffold-clusters-lorentz}).
The gain from scaffold supervision is larger under Lorentz training than under Euclidean training in every pretraining seed and at both augmentation strengths, but is primarily driven by reduced performance of the matched MolCLR baseline rather than improved scaffold-supervised models, so neither geometry consistently outperforms the other.

\begin{table}[ht]
  \caption{Molecular retrieval in the frozen embedding space $h$, after excluding molecules with the query's exact BM scaffold. Values are mean $\pm$ sample SD across three independent pretraining seeds. BM-Tan@10 compares the retrieved scaffolds, while Mol-Tan@10 compares the complete molecules. Direct ECFP4 retrieval is included as a deterministic chemical-similarity reference. Boldface marks the best learned-model mean in each column.}
  \label{tab:retrieval-main}
  \centering
  \footnotesize
  \setlength{\tabcolsep}{3.5pt}
  \begin{tabular}{@{}lcccc@{}}
    \toprule
    Model & \multicolumn{2}{c}{BM-Tan@10 $\uparrow$} & \multicolumn{2}{c}{Mol-Tan@10 $\uparrow$} \\
    \cmidrule(lr){2-3}\cmidrule(l){4-5}
    & 25\% & 15\% & 25\% & 15\% \\
    \midrule
    MolCLR (Euclidean) & $0.2749\pm0.0005$ & $0.2848\pm0.0012$ & $0.2118\pm0.0005$ & $0.2212\pm0.0004$ \\
    Scaffold (Euclidean) & $\mathbf{0.3015\pm0.0006}$ & $0.3195\pm0.0013$ & $0.2262\pm0.0004$ & $\mathbf{0.2341\pm0.0003}$ \\
    MolCLR (Lorentz) & $0.2619\pm0.0019$ & $0.2581\pm0.0010$ & $0.2228\pm0.0006$ & $0.2037\pm0.0019$ \\
    Scaffold (Lorentz) & $0.2959\pm0.0015$ & $\mathbf{0.3233\pm0.0014}$ & $\mathbf{0.2374\pm0.0007}$ & $0.2340\pm0.0005$ \\
    \midrule
    Direct ECFP4 retrieval & $0.3483$ & $0.3483$ & $0.3833$ & $0.3833$ \\
    \bottomrule
  \end{tabular}
\end{table}

\begin{figure*}[ht]
  \centering
  \includegraphics[width=\linewidth]{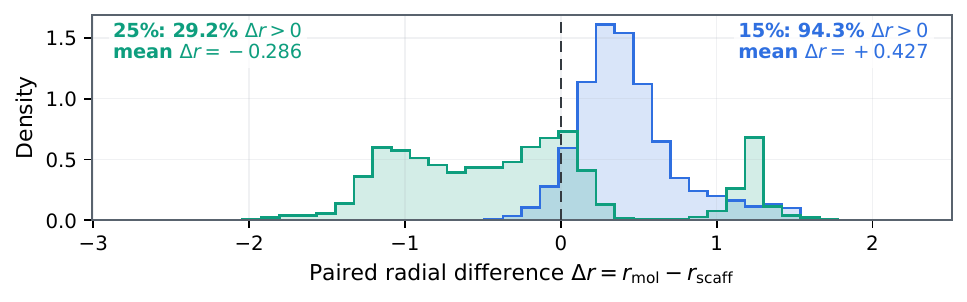}
  \caption{Paired radial difference in the bounded Lorentz alignment space $z$, for scaffold-supervised models at 15\% (blue) and 25\% (green) augmentation. Positive $\Delta r$ indicates that a molecule lies farther from the origin than its own BM scaffold. Annotations give the fraction of pairs with positive $\Delta r$ and the mean $\Delta r$ value for a single pretraining seed.}
  \label{fig:radial-difference}
\end{figure*}

\textbf{Scaffold supervision induces augmentation-dependent radial ordering.}\quad
The bounded Lorentz alignment space $z$ assigns each embedding a distance from the origin. The relative radial placement of a molecule and its own scaffold is augmentation-dependent: at 15\%, the scaffold-supervised bounded Lorentz model places the scaffold closer to the origin than its corresponding molecule for $92.0\pm2.7$\% of pairs, whereas at 25\% this occurs for only $19.3\pm8.6$\% of pairs and the mean radial ordering reverses (Figure~\ref{fig:radial-difference}). Retrieval and property prediction
are evaluated on the frozen Euclidean backbone $h$, so this ordering does not by itself
establish improved embedding quality.

\textbf{Scaffold supervision often helps property prediction.}\quad
Under Euclidean training, scaffold supervision improves the across-seed mean on six of seven tasks at 15\% augmentation and five of seven tasks at 25\% (Table~\ref{tab:frozen-plan}). The effect depends on the property: for example, ESOL improves at both Euclidean settings, whereas QM7 worsens on average. In agreement with the observed effects in the Euclidean setting, the effect of scaffold supervision under Lorentz training also helps in most tested tasks, but remains task-dependent.

\begin{table}[ht]
  \caption{Frozen linear-probe performance on seven MoleculeNet tasks. Values are mean $\pm$ SD across three independently pretrained encoders using the same downstream scaffold split. Boldface marks the best mean within each task and augmentation setting.}
  \label{tab:frozen-plan}
  \centering
  \footnotesize
  \setlength{\tabcolsep}{1.0pt}
  \resizebox{\linewidth}{!}{%
  \begin{tabular}{@{}clccccccc@{}}
    \toprule
    Aug. & Model & BBBP $\uparrow$ & BACE $\uparrow$ & ClinTox $\uparrow$ & ESOL $\downarrow$ & FreeSolv $\downarrow$ & Lipo. $\downarrow$ & QM7 $\downarrow$ \\
    \midrule
    15\% & Euc. MolCLR & $0.620\pm0.022$ & $0.635\pm0.008$ & $0.536\pm0.008$ & $1.678\pm0.094$ & $3.571\pm0.131$ & $0.997\pm0.021$ & $\mathbf{75.9\pm2.4}$ \\
    15\% & Euc. Scaffold & $0.629\pm0.003$ & $\mathbf{0.671\pm0.034}$ & $0.549\pm0.036$ & $\mathbf{1.489\pm0.027}$ & $2.940\pm0.132$ & $\mathbf{0.989\pm0.006}$ & $81.8\pm6.3$ \\
    15\% & Lor. MolCLR & $\mathbf{0.643\pm0.014}$ & $0.598\pm0.008$ & $\mathbf{0.580\pm0.034}$ & $1.668\pm0.048$ & $3.486\pm0.083$ & $1.043\pm0.007$ & $98.4\pm7.4$ \\
    15\% & Lor. Scaffold & $0.614\pm0.007$ & $0.655\pm0.068$ & $0.557\pm0.024$ & $1.573\pm0.065$ & $\mathbf{2.621\pm0.066}$ & $1.025\pm0.011$ & $96.5\pm6.9$ \\
    \midrule
    25\% & Euc. MolCLR & $0.629\pm0.008$ & $0.574\pm0.044$ & $0.502\pm0.016$ & $1.680\pm0.059$ & $3.229\pm0.049$ & $1.016\pm0.020$ & $\mathbf{76.4\pm3.0}$ \\
    25\% & Euc. Scaffold & $\mathbf{0.650\pm0.008}$ & $0.612\pm0.024$ & $0.511\pm0.037$ & $\mathbf{1.512\pm0.018}$ & $3.252\pm0.066$ & $1.000\pm0.013$ & $77.5\pm2.4$ \\
    25\% & Lor. MolCLR & $0.638\pm0.004$ & $0.658\pm0.015$ & $0.526\pm0.014$ & $1.532\pm0.021$ & $3.425\pm0.187$ & $1.027\pm0.009$ & $86.3\pm0.4$ \\
    25\% & Lor. Scaffold & $0.636\pm0.013$ & $\mathbf{0.664\pm0.029}$ & $\mathbf{0.567\pm0.017}$ & $1.565\pm0.059$ & $\mathbf{2.950\pm0.264}$ & $1.001\pm0.012$ & $86.9\pm7.8$ \\
    \bottomrule
  \end{tabular}
  }
\end{table}

\section{Limitations}

\textbf{Positioning within structure-aware molecular SSL.}\quad
This study is a controlled analysis of scaffold supervision as an additional hierarchical signal rather than a comprehensive benchmark of all prior structure-aware molecular SSL. Our results establish the effect of BM scaffold supervision relative to matched MolCLR retrainings and do not claim superiority over broader structure-aware approaches. We consider only Bemis--Murcko scaffolds, while alternative scaffold definitions or other shared structural abstractions may induce different organization. Matched comparisons with related approaches remain important future work.

\textbf{Dataset size and Lorentz ablations.}\quad
In this study, pretraining is restricted to a 1M-molecule subset. Thus, the magnitude of the observed effects may change at larger scale. Also, no scaffold-frequency balancing or cap is applied.
The bounded Lorentz implementation further uses a single $r_{\max}$, which was not independently swept.

\section{Conclusion}

We studied what happens when molecular scaffolds are used directly to supervise molecular
contrastive learning, and whether our findings are affected by embedding geometry. Scaffold supervision consistently changes how molecules are organized in the learned embedding space: molecules become closer not only to others sharing the same BM scaffold, but also to molecules with related, nonidentical scaffolds. The same embeddings improve property prediction on several tasks, although these gains depend on the property being predicted.
Embedding geometry changes the strength of the scaffold-supervision effect, but Lorentz training does not provide a consistent advantage over Euclidean training. Together, our results show that explicitly teaching a model the relation between a molecule and its structural core provides a simple way to shape molecular embeddings, while the usefulness of this organization depends on the downstream task.

\section{Acknowledgements}

Research reported in this work was partially or completely facilitated by computational resources and support of the Delft AI Cluster (DAIC) at TU Delft (RRID: SCR\_025091), but remains the sole responsibility of the authors, not the DAIC team \citep{DAIC}.

\clearpage
\bibliographystyle{plainnat}
\bibliography{references}

\appendix

\section{Objectives and Geometry}
\label{sec:geometry}
For a minibatch of $N$ molecules, two augmented views yield $2N$
embeddings. For a positive pair $(z_i,z_j)$ from the same molecule, NT-Xent
\citep{SimCLR} is
\begin{equation}
 \mathcal L_{\rm mol}=-\log
 \frac{\exp\!\big(\mathrm{sim}(z_i,z_j)/\tau\big)}
 {\sum_{k=1}^{2N}\mathbf 1_{[k\ne i]}
  \exp\!\big(\mathrm{sim}(z_i,z_k)/\tau\big)}
 \label{eq:ntxent}
\end{equation}
where $\mathrm{sim}$ is the geometry specific score and $\tau$ is a temperature parameter. 

For explicit alignment, let $U$ be the number of distinct nonempty scaffolds
in the batch, let $z_{S_u}$ be the projection of scaffold $u$, and let $S(G)$
index the scaffold of molecule $G$. The loss is
\begin{equation}
 \mathcal L_{\rm scaff}
 =-\frac{1}{2}\,\mathbb E_G\sum_{v\in\{i,j\}}
 \log\frac{\exp\!\big(\mathrm{sim}(z_{G^{(v)}},z_{S(G)})/\tau_s\big)}
 {\sum_{u=1}^{U}\exp\!\big(\mathrm{sim}(z_{G^{(v)}},z_{S_u})/\tau_s\big)}
 \label{eq:scaffold}
\end{equation}
where the expectation is over molecules with nonempty scaffolds. 
The total objective is $\mathcal L=\mathcal L_{\rm mol}+\lambda_s\mathcal L_{\rm scaff}$.

\paragraph{Primer on the Lorentz model of hyperbolic space.}
Hyperbolic space has constant negative curvature, and the volume available at
radius $r$ grows exponentially rather than polynomially. This makes it a
natural candidate geometry for tree-like data, but whether that inductive bias
helps molecular embeddings is an empirical question
\citep{Poincare,Sarkar2011,LorentzEmbeddings}.

A smooth $n$-dimensional manifold $\mathcal M$ is locally parametrized by
$\mathbb R^n$. A Riemannian metric $\mathfrak g$ assigns to every
$x\in\mathcal M$ a positive-definite inner product $\mathfrak g_x$ on the
tangent space $T_x\mathcal M$, varying smoothly with $x$. The pair
$(\mathcal M,\mathfrak g)$ is a Riemannian manifold; its metric defines lengths
of tangent vectors and curves, and hence geodesic distance.

The Lorentz model represents an $n$-dimensional hyperbolic space as the upper
sheet of a hyperboloid in $\mathbb R^{n+1}$. Writing
$x=[x_t,x_s]$ and $y=[y_t,y_s]$, with one time coordinate and $n$ spatial
coordinates, its inner product is
\[
 \langle x,y\rangle_{\mathcal L}=-x_ty_t+x_s^\top y_s
\]
For curvature $-c$, with $c>0$, the manifold is
\[
 \mathcal L_c^n=
 \left\{x\in\mathbb R^{n+1}:\langle x,x\rangle_{\mathcal L}=-\frac1c,
 \ x_t>0\right\}
\]
and its origin is $O=[1/\sqrt c,\mathbf 0]$. The geodesic distance between
$x,y\in\mathcal L_c^n$ is
\begin{equation}
 d_{\mathcal L}(x,y)=\frac1{\sqrt c}
 \mathrm{arccosh}\!\left(-c\langle x,y\rangle_{\mathcal L}\right)
 \label{eq:lorentz}
\end{equation}

The tangent space $T_x\mathcal M$ is the $n$-dimensional vector space that 
linearly approximates the manifold in $x$. For the Lorentz hyperboloid,
$T_x\mathcal L_c^n=\{w\in\mathbb R^{n+1}:\langle x,w\rangle_{\mathcal L}=0\}$.
At $O=[1/\sqrt c,\mathbf 0]$, this condition gives $w_t=0$, so
$T_O\mathcal L_c^n$ is naturally identified with $\mathbb R^n$ through
$v\mapsto[0,v]$. The implementation scales $z$ in this tangent space and maps
it to the hyperboloid with the exponential map.

The Lorentz model uses $-d_{\mathcal L}$ as its contrastive score. Following
the implementation, a scaled projection $v=\alpha z\in\mathbb R^{256}$ is
lifted from the tangent space at $O$ with
\[
 \exp_O(v)=\left[
 \frac{\cosh(\sqrt c\|v\|)}{\sqrt c},\
 \frac{\sinh(\sqrt c\|v\|)}{\sqrt c\|v\|}v
 \right]\in\mathbb R^{257}
\]
using the continuous value at $v=0$. Thus the hyperboloid has intrinsic
dimension 256 and ambient dimension 257. The Lorentz variant substitutes
$-d_{\mathcal L}\!\left(\exp_O(\alpha z),\exp_O(\alpha z')\right)$ for cosine
similarity. No manifold map is applied to the backbone embedding $h$.

\paragraph{Bounded and unbounded Lorentz formulations.}
We consider two implementations of the Lorentz objective, differing only in how the scaled projection is mapped onto the hyperboloid. The \emph{unbounded} formulation follows the mapping above, lifting $v=\alpha z$ directly with $\exp_O$. The \emph{bounded} formulation instead replaces $v$ before the exponential map by
\[
 \widetilde v=
 r_{\max}\tanh\!\left(\frac{\|v\|}{r_{\max}}\right)
 \frac{v}{\|v\|}   
\]
with the continuous value at $v=0$.

\section{Experimental Details and Runtimes}
\label{sec:experimental-details}

For downstream evaluation, each pretrained encoder is probed with three linear heads whose scores are first averaged within the encoder, to produce per-seed downstream results. We then report the mean and sample standard deviation across three independently pretrained encoders. The backbone is frozen throughout.

Pretraining used one NVIDIA A40 GPU per run. Representative runs
required approximately 4.8 minutes per epoch for Euclidean MolCLR and
8.3 minutes per epoch with scaffold alignment, corresponding to approximately
4.0 and 6.9 hours, respectively, for 50 epochs. The corresponding Lorentz runs required approximately 4.5 minutes per epoch for MolCLR and
8.2 minutes per epoch with scaffold alignment, or approximately 3.8 and
6.8 hours for 50 epochs. Downstream linear-probe experiments were also 
run on NVIDIA A40 GPUs. Configuration details are reported 
in Table ~\ref{tab:reproducibility}.

\begin{table}[ht]
 \caption{Pretraining and frozen-probe configuration.}
 \label{tab:reproducibility}
 \centering
 \footnotesize
 \setlength{\tabcolsep}{3pt}
 \begin{tabular}{@{}p{0.22\linewidth}p{0.72\linewidth}@{}}
  \toprule
  Component & Final specification \\
  \midrule
  Architecture & Five-layer shared GINE, hidden width 300 and mean
  pooling; $h\in\mathbb R^{512}$ and $z\in\mathbb R^{256}$. \\
  Views & Random atom-feature masking and bond deletion,
  at 15\% or 25\%. \\
  Optimization & Adam, batch 512, 50 epochs, initial learning rate
  $5\!\times\!10^{-4}$, weight decay $10^{-5}$, 10 warm-up epochs followed by
  cosine decay. \\
  Loss and geometry & $\tau=\tau_s=0.1$, $\lambda_s=0.1$, fixed $c=1$, and a
  learned Lorentz scale $\alpha$. For bounded Lorentz, $r_{\max}=5$. \\
  Frozen probes & One deterministic 80/10/10 BM-scaffold split; three independently
  pretrained encoders per condition and three linear heads per encoder on frozen
  $h$, batch 32, 100 epochs, Adam at $10^{-3}$ with weight decay $10^{-6}$. \\
  \bottomrule
 \end{tabular}
\end{table}

\textbf{Downstream task definitions.}\quad The benchmarks span distinct types of molecular properties. BBBP, BACE, and ClinTox probe biological or physiological outcomes, whereas ESOL, FreeSolv, and Lipophilicity describe physicochemical properties. QM7 instead targets a quantum-mechanical energetic property. Consequently, scaffold information need not be equally predictive across all seven tasks. The selected tasks are reported in Table ~\ref{tab:moleculenet-tasks}.

\begin{table}[ht]
\centering
\small
\caption{Downstream MoleculeNet benchmarks used for frozen linear probing.
Classification tasks are evaluated by ROC-AUC (higher is better), while regression
tasks use RMSE or MAE (lower is better). All experiments use the scaffold splits
described in the main text.}
\label{tab:moleculenet-tasks}
\begin{tabular}{llll}
\toprule
Dataset & Prediction target & Task & Metric \\
\midrule
BBBP & Blood--brain barrier penetration & Classification & ROC-AUC $\uparrow$ \\

BACE & BACE-1 inhibitor activity & Classification & ROC-AUC $\uparrow$ \\

ClinTox & Clinical toxicity / FDA approval & Classification & ROC-AUC $\uparrow$ \\

ESOL & Aqueous solubility & Regression & RMSE $\downarrow$ \\

FreeSolv & Hydration free energy in water & Regression & RMSE $\downarrow$ \\

Lipophilicity & Octanol/water distribution ($\log D$, pH 7.4) & Regression & RMSE $\downarrow$ \\

QM7 & Atomization energy & Regression & MAE $\downarrow$ \\
\bottomrule
\end{tabular}
\end{table}

\section{Complete Downstream Results}
\label{sec:complete-downstream}
Classification uses ROC--AUC ($\uparrow$); ESOL, FreeSolv,
and Lipophilicity use RMSE ($\downarrow$); QM7 uses MAE ($\downarrow$).
Here we also provide the unbounded Lorentz rows for comparison.

\begin{table}[ht]
  \caption{Downstream results at 15\% augmentation, with the same aggregation 
  and metrics as Table~\ref{tab:frozen-full}.
  Boldface and underlining mark the best and second-best models in each column.}
  \label{tab:frozen-15}
  \centering
  \tiny
  \setlength{\tabcolsep}{0.8pt}
  \resizebox{\linewidth}{!}{%
  \begin{tabular}{@{}lccccccc@{}}
    \toprule
    Model & BBBP $\uparrow$ & BACE $\uparrow$ & ClinTox $\uparrow$ & ESOL $\downarrow$ & FreeSolv $\downarrow$ & Lipo. $\downarrow$ & QM7 $\downarrow$ \\
    \midrule
    MolCLR (Euclidean) & $0.620\!\pm\!0.022$ & $0.635\!\pm\!0.008$ & $0.536\!\pm\!0.008$ & $1.678\!\pm\!0.094$ & $3.571\!\pm\!0.131$ & $\underline{0.997\!\pm\!0.021}$ & $\mathbf{75.93\!\pm\!2.40}$ \\
    Scaffold (Euclidean) & $\underline{0.629\!\pm\!0.003}$ & $\mathbf{0.671\!\pm\!0.034}$ & $0.549\!\pm\!0.036$ & $\mathbf{1.489\!\pm\!0.027}$ & $\underline{2.940\!\pm\!0.132}$ & $\mathbf{0.989\!\pm\!0.006}$ & $\underline{81.83\!\pm\!6.32}$ \\
    MolCLR (Lorentz--unbounded) & $0.627\!\pm\!0.020$ & $0.598\!\pm\!0.008$ & $0.506\!\pm\!0.016$ & $1.715\!\pm\!0.025$ & $3.530\!\pm\!0.188$ & $1.046\!\pm\!0.008$ & $101.28\!\pm\!2.37$ \\
    Scaffold (Lorentz--unbounded) & $0.613\!\pm\!0.005$ & $0.641\!\pm\!0.109$ & $0.550\!\pm\!0.028$ & $1.674\!\pm\!0.040$ & $3.098\!\pm\!0.262$ & $1.010\!\pm\!0.017$ & $86.72\!\pm\!8.86$ \\
    MolCLR (Lorentz--bounded) & $\mathbf{0.643\!\pm\!0.014}$ & $0.598\!\pm\!0.008$ & $\mathbf{0.580\!\pm\!0.034}$ & $1.668\!\pm\!0.048$ & $3.486\!\pm\!0.083$ & $1.043\!\pm\!0.007$ & $98.41\!\pm\!7.37$ \\
    Scaffold (Lorentz--bounded) & $0.614\!\pm\!0.007$ & $\underline{0.655\!\pm\!0.068}$ & $\underline{0.557\!\pm\!0.024}$ & $\underline{1.573\!\pm\!0.065}$ & $\mathbf{2.621\!\pm\!0.066}$ & $1.025\!\pm\!0.011$ & $96.47\!\pm\!6.89$ \\
    \bottomrule
  \end{tabular}
  }
\end{table}

\begin{table}[ht]
  \caption{Downstream results at 25\% augmentation.
  Boldface and underlining mark the best and second-best models in each column.}
  \label{tab:frozen-full}
  \centering
  \tiny
  \setlength{\tabcolsep}{0.8pt}
  \resizebox{\linewidth}{!}{%
  \begin{tabular}{@{}lccccccc@{}}
    \toprule
    Model & BBBP $\uparrow$ & BACE $\uparrow$ & ClinTox $\uparrow$ & ESOL $\downarrow$ & FreeSolv $\downarrow$ & Lipo. $\downarrow$ & QM7 $\downarrow$ \\
    \midrule
    MolCLR (Euclidean) & $0.629\!\pm\!0.008$ & $0.574\!\pm\!0.044$ & $0.502\!\pm\!0.016$ & $1.680\!\pm\!0.059$ & $3.229\!\pm\!0.049$ & $1.016\!\pm\!0.020$ & $\mathbf{76.37\!\pm\!3.04}$ \\
    Scaffold (Euclidean) & $\mathbf{0.650\!\pm\!0.008}$ & $0.612\!\pm\!0.024$ & $0.511\!\pm\!0.037$ & $\mathbf{1.512\!\pm\!0.018}$ & $3.252\!\pm\!0.066$ & $\underline{1.000\!\pm\!0.013}$ & $\underline{77.51\!\pm\!2.41}$ \\
    MolCLR (Lorentz--unbounded) & $0.633\!\pm\!0.011$ & $0.651\!\pm\!0.012$ & $0.503\!\pm\!0.013$ & $\underline{1.517\!\pm\!0.015}$ & $\mathbf{2.708\!\pm\!0.131}$ & $1.030\!\pm\!0.014$ & $96.20\!\pm\!9.88$ \\
    Scaffold (Lorentz--unbounded) & $0.620\!\pm\!0.030$ & $\mathbf{0.690\!\pm\!0.024}$ & $\mathbf{0.592\!\pm\!0.035}$ & $1.595\!\pm\!0.045$ & $\underline{2.741\!\pm\!0.416}$ & $\mathbf{0.993\!\pm\!0.018}$ & $81.46\!\pm\!3.95$ \\
    MolCLR (Lorentz--bounded) & $\underline{0.638\!\pm\!0.004}$ & $0.658\!\pm\!0.015$ & $0.526\!\pm\!0.014$ & $1.532\!\pm\!0.021$ & $3.425\!\pm\!0.187$ & $1.027\!\pm\!0.009$ & $86.35\!\pm\!0.40$ \\
    Scaffold (Lorentz--bounded) & $0.636\!\pm\!0.013$ & $\underline{0.664\!\pm\!0.029}$ & $\underline{0.567\!\pm\!0.017}$ & $1.565\!\pm\!0.059$ & $2.950\!\pm\!0.264$ & $1.001\!\pm\!0.012$ & $86.90\!\pm\!7.87$ \\
    \bottomrule
  \end{tabular}
  }
\end{table}

\section{Structural Evaluation}
\label{sec:structural-diagnostics}
We use one fixed cohort of 100,000 randomly selected molecules from the pretraining validation set. Molecules are extracted and retained when they: have a valid stored BM scaffold, can be reconstructed as an RDKit molecule, and are not exact-molecule duplicates; 
no scaffold-frequency balancing or cap is applied. 
The same cohort is used for all evaluated encoders.

The cohort contains 57,939 exact BM scaffolds. Of the 100,000 queries, 48,762
belong to 6,701 multi-member scaffold groups and therefore have at least one
exact-BM match. The remaining 51,238 have singleton exact scaffolds. Each query
is removed from its own candidate set. Exact-BM evaluation is restricted to
queries with at least one other exact match. Graded BM-scaffold evaluation uses every
query, removes all candidates with the query's exact BM scaffold, and scores
the remaining neighbours by ECFP4--Tanimoto similarity between the query and
candidate BM scaffolds.

For the exact-BM retrieval, MAP@$k$ averages precision at relevant ranks and
normalizes by the smaller number between $k$ and the amount of available positives.
BM-Tan@$k$ is the mean BM-scaffold ECFP4 similarity among the top $k$, and
nDCG@10 uses that similarity as a relevance score. Table~\ref{tab:retrieval-h} reports retrieval in the backbone space $h$.

\begin{table}[ht]
  \caption{Retrieval results in backbone space $h$.
  Learned-model values are mean $\pm$ sample SD across three independently
  pretrained encoders. Exact BM is an objective-proximal check; graded metrics
  exclude every candidate with the query's exact BM scaffold. Bold and
  underlining mark the best and second-best learned-model point estimates
  within each augmentation setting. 
  Full-molecule ECFP4 is a deterministic, chemical-similarity reference.}
  \label{tab:retrieval-h}
  \centering
  \scriptsize
  \setlength{\tabcolsep}{2.5pt}
  \begin{tabular}{@{}clccc@{}}
    \toprule
    Aug. & Model & Exact MAP@10 $\uparrow$  & BM-Tan@10 $\uparrow$  & BM-nDCG@10 $\uparrow$  \\
    \midrule
    15\% & MolCLR (Euclidean) & $0.2904\pm0.0025$ & $0.2848\pm0.0012$ & $0.5169\pm0.0019$ \\
    15\% & Scaffold (Euclidean) & $0.4068\pm0.0024$ & $0.3195\pm0.0013$ & $\underline{0.5738\pm0.0021}$ \\
    15\% & MolCLR (Lorentz--unbounded) & $0.2208\pm0.0014$ & $0.2492\pm0.0004$ & $0.4483\pm0.0011$ \\
    15\% & Scaffold (Lorentz--unbounded) & $\underline{0.4233\pm0.0100}$ & $\underline{0.3223\pm0.0034}$ & $\mathbf{0.5740\pm0.0068}$ \\
    15\% & MolCLR (Lorentz--bounded) & $0.2375\pm0.0026$ & $0.2581\pm0.0010$ & $0.4629\pm0.0015$ \\
    15\% & Scaffold (Lorentz--bounded) & $\mathbf{0.4407\pm0.0120}$ & $\mathbf{0.3233\pm0.0014}$ & $0.5721\pm0.0023$ \\
    \midrule
    25\% & MolCLR (Euclidean) & $0.2635\pm0.0025$ & $0.2749\pm0.0005$ & $0.4994\pm0.0010$ \\
    25\% & Scaffold (Euclidean) & $\underline{0.3264\pm0.0030}$ & $\mathbf{0.3015\pm0.0006}$ & $\mathbf{0.5460\pm0.0007}$ \\
    25\% & MolCLR (Lorentz--unbounded) & $0.1670\pm0.0045$ & $0.2516\pm0.0029$ & $0.4434\pm0.0053$ \\
    25\% & Scaffold (Lorentz--unbounded) & $\mathbf{0.3288\pm0.0060}$ & $\underline{0.2992\pm0.0010}$ & $\underline{0.5270\pm0.0018}$ \\
    25\% & MolCLR (Lorentz--bounded) & $0.2090\pm0.0085$ & $0.2619\pm0.0019$ & $0.4613\pm0.0031$ \\
    25\% & Scaffold (Lorentz--bounded) & $0.3150\pm0.0138$ & $0.2959\pm0.0015$ & $0.5211\pm0.0025$ \\
    \midrule
    -- & Full-molecule ECFP4/Tanimoto & 0.2601 & 0.3483 & 0.6267 \\
    \bottomrule
  \end{tabular}
\end{table}

\begin{table}[ht]
  \caption{Full-molecule similarity of neighbours selected in backbone space
  $h$, after excluding exact-BM matches. Values are mean $\pm$ sample SD across
  three independently pretrained encoders. Boldface and underlining mark the
  best and second-best learned-model point estimates within each augmentation
  setting.}
  \label{tab:full-molecule-retrieval}
  \centering
  \scriptsize
  \begin{tabular}{@{}clcc@{}}
    \toprule
    Aug. & Model & Mol-Tan@10 $\uparrow$  & Mol-nDCG@10 $\uparrow$ \\
    \midrule
    15\% & MolCLR (Euclidean) & $0.2212\pm0.0004$ & $0.5680\pm0.0009$ \\
    15\% & Scaffold (Euclidean) & $\mathbf{0.2341\pm0.0003}$ & $\mathbf{0.6015\pm0.0008}$ \\
    15\% & MolCLR (Lorentz--unbounded) & $0.1952\pm0.0006$ & $0.4982\pm0.0017$ \\
    15\% & Scaffold (Lorentz--unbounded) & $0.2332\pm0.0011$ & $0.5989\pm0.0030$ \\
    15\% & MolCLR (Lorentz--bounded) & $0.2037\pm0.0019$ & $0.5206\pm0.0050$ \\
    15\% & Scaffold (Lorentz--bounded) & $\underline{0.2340\pm0.0005}$ & $\underline{0.6007\pm0.0013}$ \\
    \midrule
    25\% & MolCLR (Euclidean) & $0.2118\pm0.0005$ & $0.5418\pm0.0016$ \\
    25\% & Scaffold (Euclidean) & $0.2262\pm0.0004$ & $0.5808\pm0.0011$ \\
    25\% & MolCLR (Lorentz--unbounded) & $0.2178\pm0.0028$ & $0.5591\pm0.0075$ \\
    25\% & Scaffold (Lorentz--unbounded) & $\mathbf{0.2392\pm0.0016}$ & $\mathbf{0.6147\pm0.0044}$ \\
    25\% & MolCLR (Lorentz--bounded) & $0.2228\pm0.0006$ & $0.5720\pm0.0016$ \\
    25\% & Scaffold (Lorentz--bounded) & $\underline{0.2374\pm0.0007}$ & $\underline{0.6097\pm0.0020}$ \\
    \midrule
    -- & Direct ECFP4 & $0.3833$ & $1.0000$ \\
    \bottomrule
  \end{tabular}
\end{table}

\section{Projection-space geometry diagnostics}
\label{sec:diagnostics}
The bounded Lorentz formulation used in the main experiments was adopted after examining the original unbounded implementation. There, every embedding in the alignment space $z$ ends up at essentially the same distance from the origin, as their mapped radii reach the numerical stability limit of the hyperbolic implementation. Because all points share that radius, negative geodesic distance becomes almost perfectly rank-equivalent to raw-$z$ cosine similarity 
(Spearman correlation $\approx$ 1.0). Thus these models effectively lose radial variation, although the resulting contrastive objective is not identical to cosine NT-Xent because the transformed logits and gradients still differ. 
The bounded variant prevents embeddings from reaching this numerical limit and restores variation in distance from the origin. 
These diagnostics concern $z$, on which the Lorentz objective operates, while retrieval and downstream evaluation use the frozen Euclidean backbone $h$.

\section{Augmentation Diagnostics}
\label{sec:augmentation-diagnostics}

We characterize how augmentation strength affects molecular graph connectivity
independently of the learned embeddings. For each operator and perturbation
strength, we generate 18,000 stochastic views from a fixed set of 3,000
validation molecules. We measure the number of connected components among
non-isolated atoms, the fraction of atoms retained in the largest connected
component, and the fraction of views that split into at least two multi-atom
components.

\begin{figure}[ht]
  \centering
  \includegraphics[width=\linewidth]{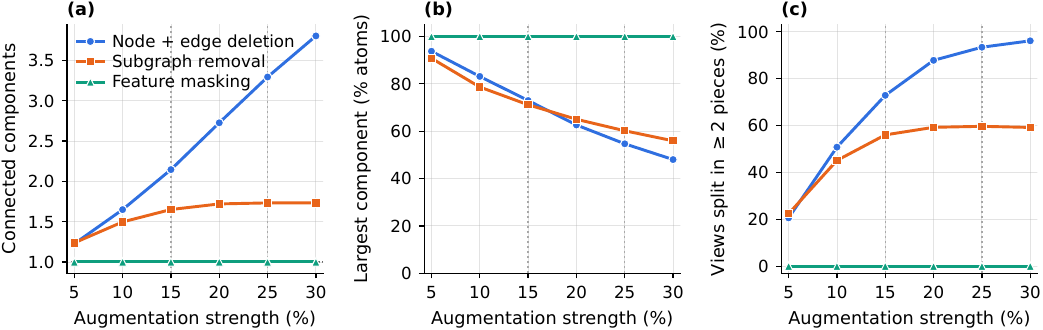}
  \caption{Effect of augmentation strength on molecular graph connectivity.
  Dashed vertical lines indicate the 15\% and 25\% settings used in the main
  experiments. $\textit{(a):}$ Mean number of connected components among non-isolated atoms.
  $\textit{(b):}$ Fraction of atoms contained in the largest connected component.
  $\textit{(c):}$ Fraction of augmented views split into at least two multi-atom components.
  Each point aggregates 18,000 stochastic views; feature masking preserves
  graph topology by construction.}
  \label{fig:fragmentation-sweep}
\end{figure}

Figure~\ref{fig:fragmentation-sweep} shows these quantities across perturbation
strengths from 5\% to 30\%. For the augmentation used in our reported models,
connectivity degrades progressively with increasing perturbation strength:
15\% and 25\% therefore correspond to milder and stronger points along the same
fragmentation regime rather than qualitatively distinct perturbations.
We add subgraph removal and feature masking as structural controls.
Subgraph removal produces a different fragmentation profile because it removes
a localized region instead of randomly dropping bonds. Feature masking preserves graph 
connectivity by construction, by masking bond information instead of removing it.

These diagnostics quantify structural fragmentation. Fragmented augmented views are not removed or re-sampled, they are passed to the encoder as disconnected graphs, even though they no longer correspond to a chemically valid single molecule.

\section{Backbone-Space Visualization}
\label{sec:embedding-maps}

To qualitatively illustrate the structural organization measured by retrieval, we visualize the frozen backbone embedding $h$ of matched MolCLR and scaffold-supervised models at 15\% augmentation, under both Euclidean and Lorentz training. We use the same molecule subset and t-SNE configuration for all four models, and compute the embedding from cosine distances between L2-normalized backbone embeddings.

\begin{figure*}[ht]
  \centering
  \begin{minipage}{0.48\textwidth}
    \centering
    \includegraphics[width=\linewidth]{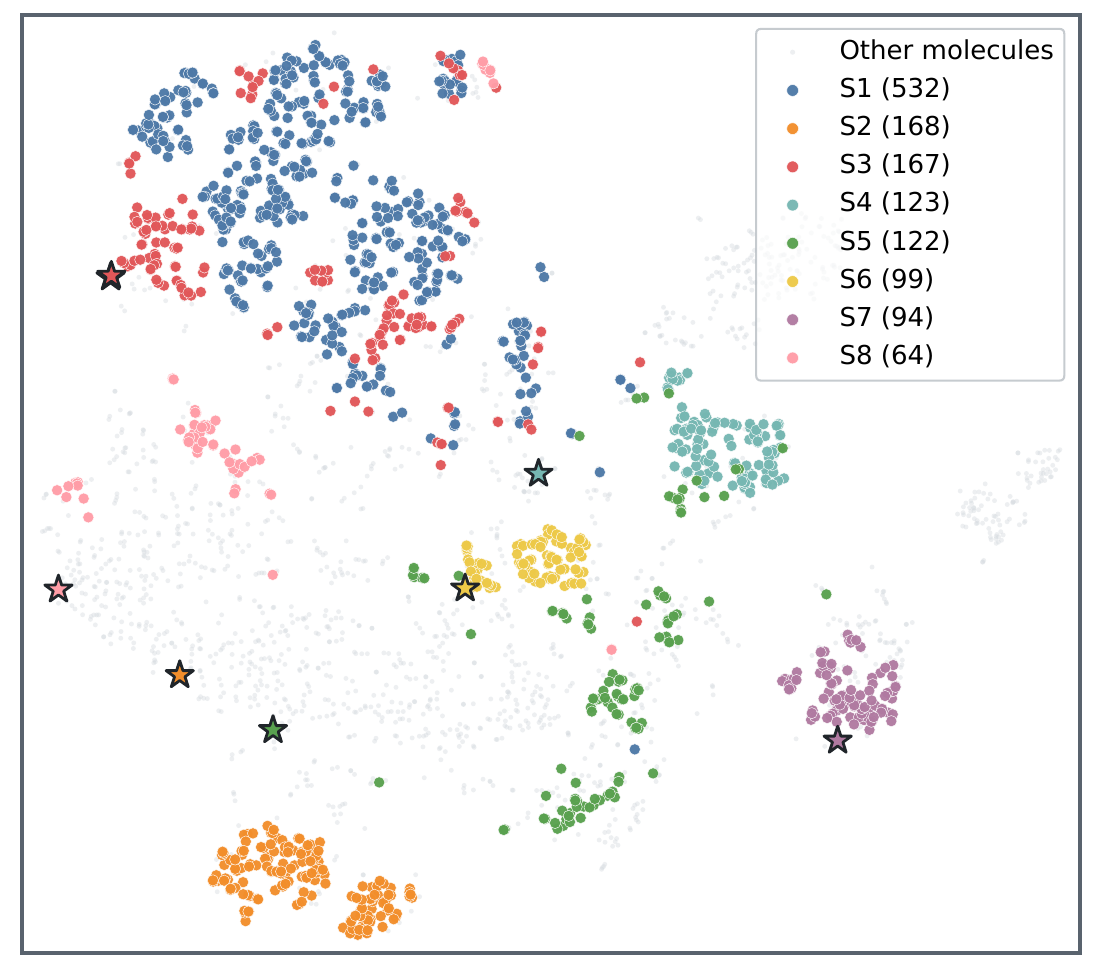}\\[-1pt]
    {\small (a) MolCLR - Euclidean}
  \end{minipage}\hfill
  \begin{minipage}{0.48\textwidth}
    \centering
    \includegraphics[width=\linewidth]{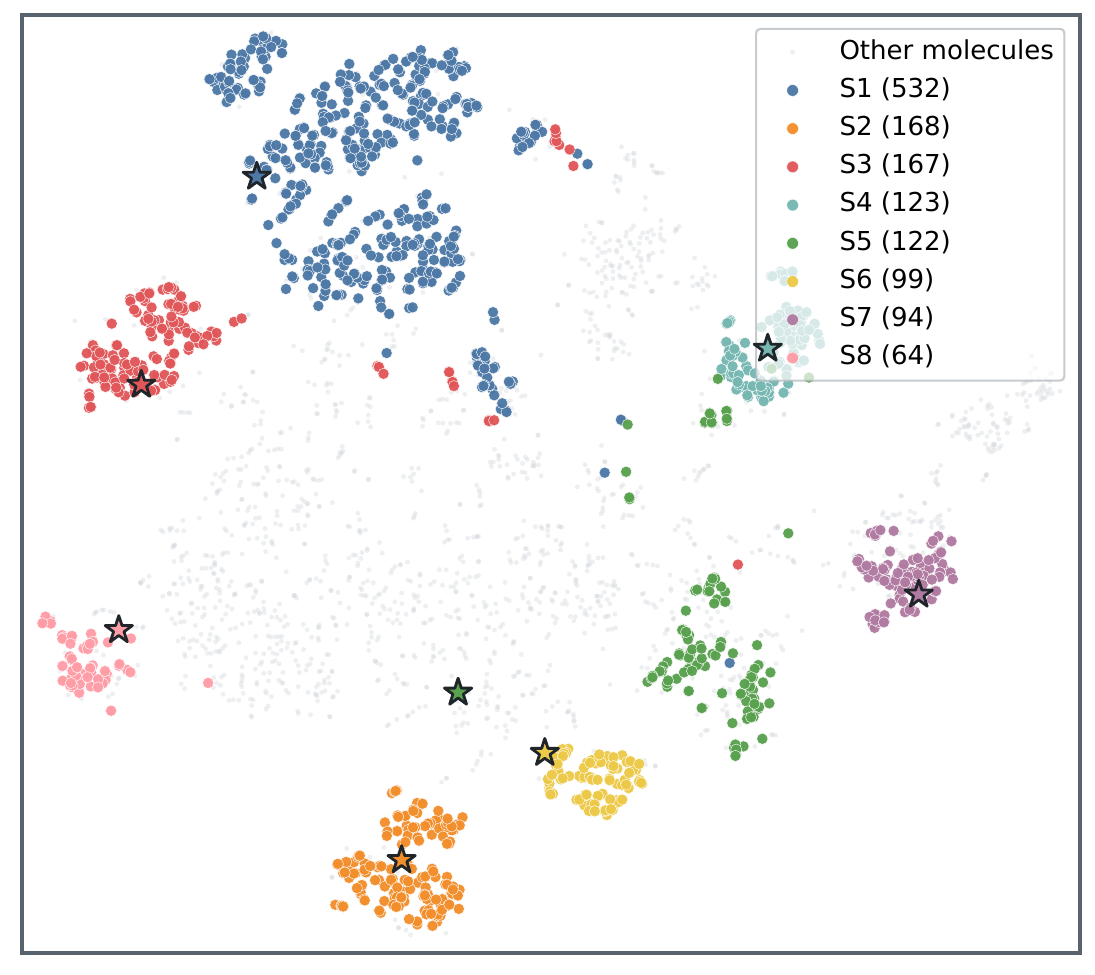}\\[-1pt]
    {\small (b) Scaffold supervision - Euclidean}
  \end{minipage}

  \caption{Qualitative visualization of backbone-space $h$ organization at 15\%
  augmentation for one Euclidean model. Colors identify molecules belonging to eight frequent
  Bemis--Murcko scaffolds in that set, while grey points show additional molecules 
  from the shared evaluation subset. Stars indicate the embeddings of the
  corresponding scaffold graphs. Both panels use the same molecules and t-SNE
  settings.}
  \label{fig:embedding-scaffold-clusters}
\end{figure*}

\begin{figure*}[ht]
  \centering
  \begin{minipage}{0.48\textwidth}
    \centering
    \includegraphics[width=\linewidth]{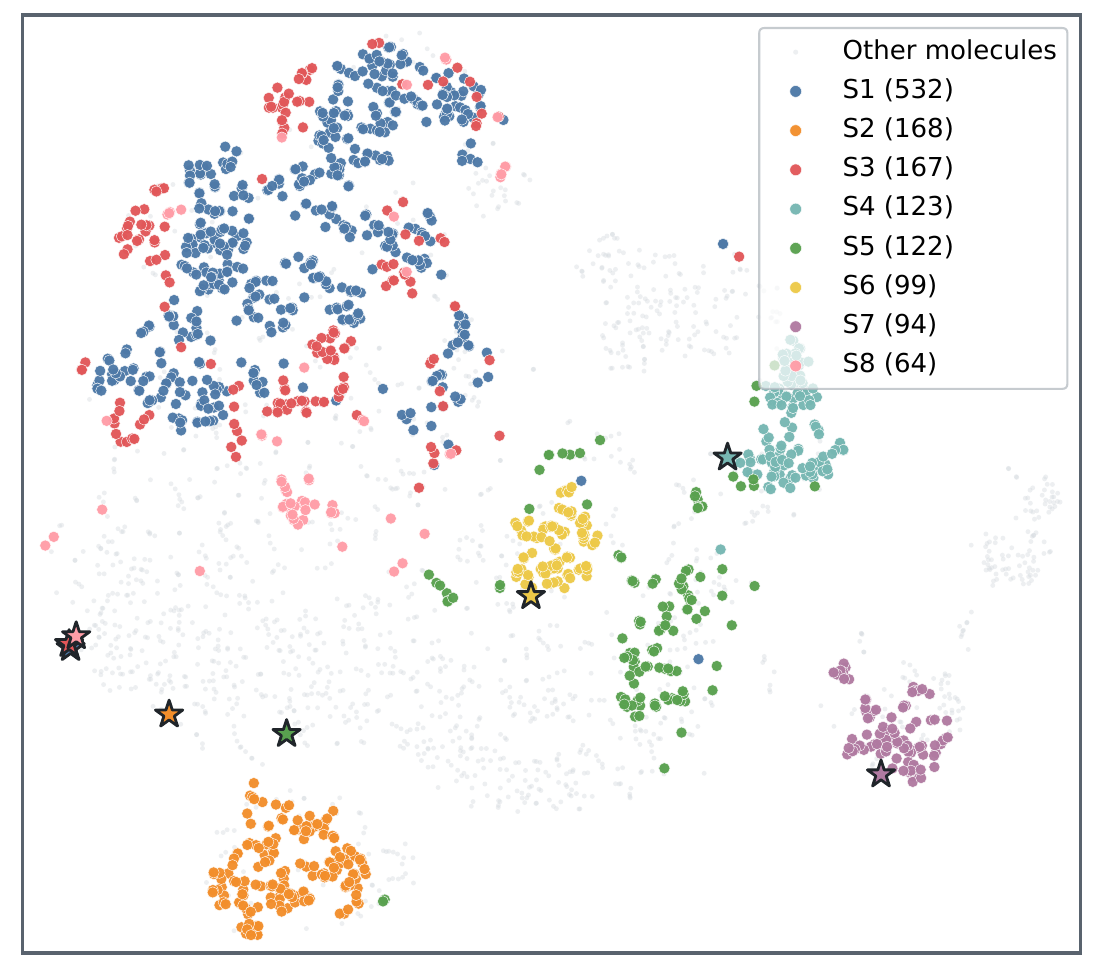}\\[-1pt]
    {\small (a) MolCLR - Lorentz}
  \end{minipage}\hfill
  \begin{minipage}{0.48\textwidth}
    \centering
    \includegraphics[width=\linewidth]{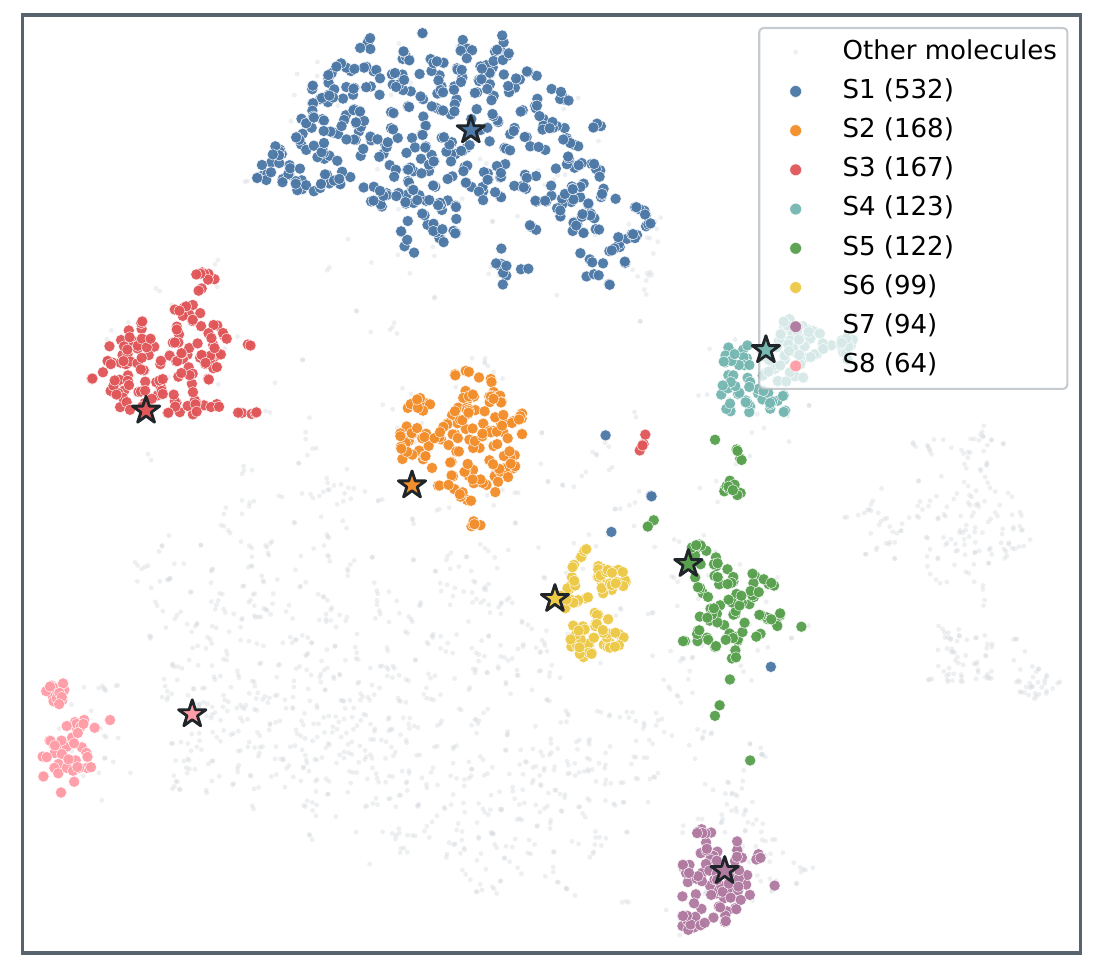}\\[-1pt]
    {\small (b) Scaffold supervision - Lorentz}
  \end{minipage}

  \caption{Qualitative visualization of backbone-space $h$ organization at 15\%
  augmentation for one bounded Lorentz model. Colors identify molecules belonging to eight frequent
  Bemis--Murcko scaffolds in that set, while grey points show additional molecules 
  from the shared evaluation subset. Stars indicate the embeddings of the
  corresponding scaffold graphs. Both panels use the same molecules and t-SNE
  settings.}
  \label{fig:embedding-scaffold-clusters-lorentz}
\end{figure*}

\begin{figure*}[ht]
  \centering
  \includegraphics[width=\linewidth]{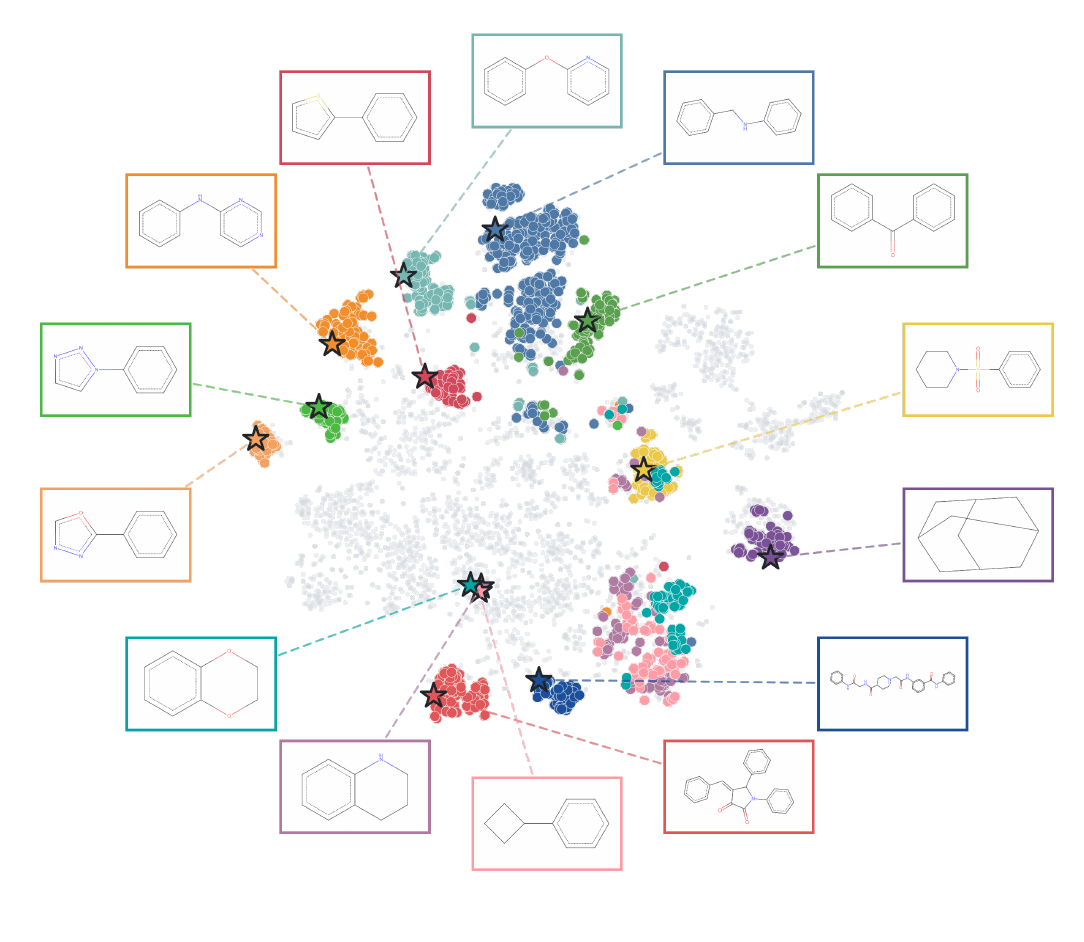}
  \caption{Qualitative visualization of the scaffold supervised
  backbone space $h$ at 15\% augmentation. Here we show 6000 molecules
  and 14 frequent Bemis--Murcko scaffolds. Colored points
  denote molecules assigned to each scaffold, grey points provide molecular
  context, and stars mark the corresponding scaffold embeddings.}
  \label{fig:embedding-scaffold-expanded}
\end{figure*}

Figures ~\ref{fig:embedding-scaffold-clusters} and ~\ref{fig:embedding-scaffold-clusters-lorentz} show more visibly coherent
scaffold-specific groups under scaffold supervision for the Euclidean and Lorentz implementations, respectively, consistent with the quantitative retrieval results in
Table~\ref{tab:retrieval-h}. In particular, several scaffold groups that overlap
under MolCLR become more clearly separated after scaffold supervision, while
the corresponding scaffold embeddings tend to lie closer to their
associated molecular groups. Figure~\ref{fig:embedding-scaffold-expanded} provides a more comprehensive view of how different scaffold groups are distributed in the scaffold-supervised backbone for the Euclidean model.

\clearpage

\end{document}